\documentclass[conference, letterpaper]{IEEEtran}

\IEEEoverridecommandlockouts                              

\usepackage[font=small]{caption}
\usepackage{amssymb}
\usepackage{graphicx}
\usepackage{mathtools}
\usepackage{amsmath}
\usepackage{gensymb}
\usepackage{float}
\usepackage{multirow}
\usepackage{makecell}
\usepackage{mathtools}
\usepackage{siunitx}
\usepackage{overpic}
\usepackage{subcaption}
\usepackage{amsmath}
\usepackage{titlesec}

\usepackage{tabularx}
    \newcolumntype{L}{>{\raggedright\arraybackslash}X}
\usepackage[utf8]{inputenc}
\usepackage[english]{babel}

\usepackage{amsthm}
\usepackage{booktabs}
\usepackage{algorithm}
\usepackage{algpseudocode}
\usepackage[textsize=tiny]{todonotes}

\usepackage{subcaption}

\title{\LARGE \bf \raisebox{-0.3\height}{\includegraphics[height=3.5em]{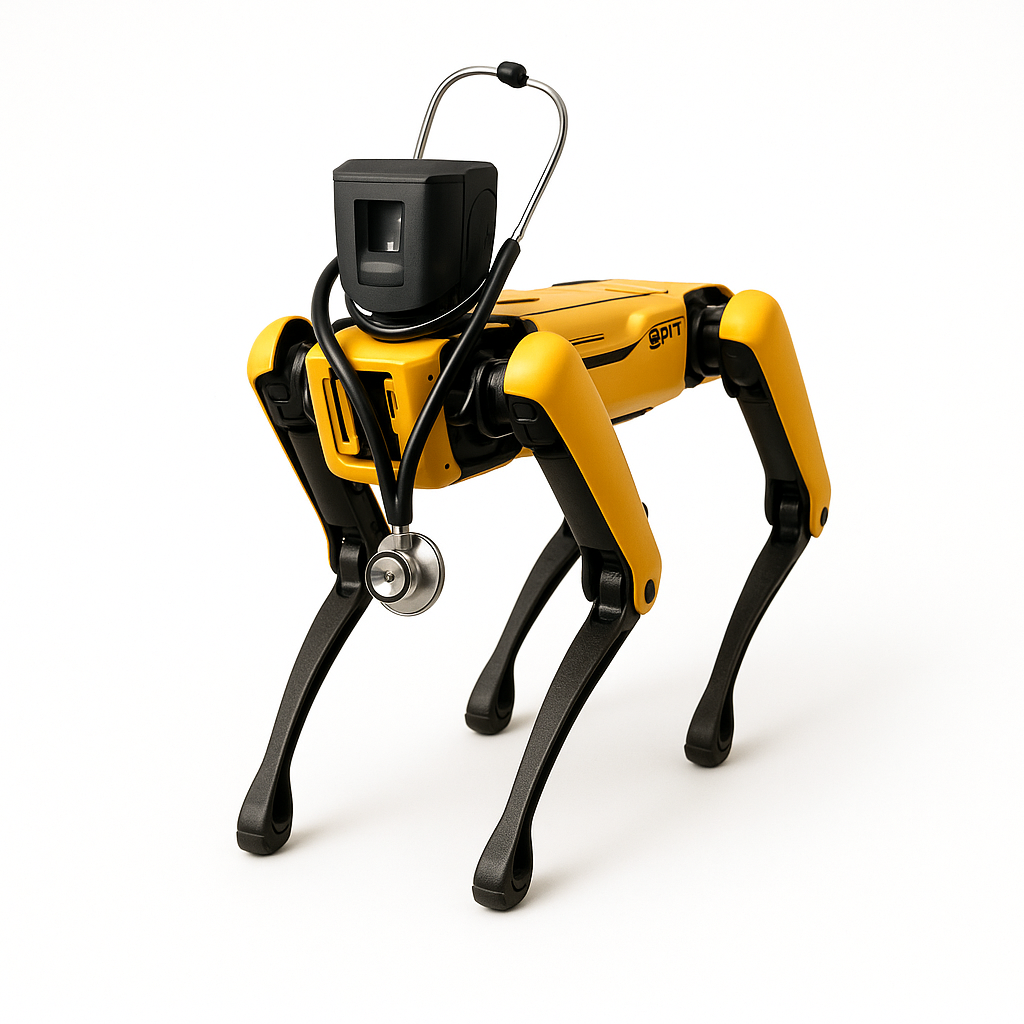}}DR. Nav: Semantic-Geometric Representations for Proactive Dead-End Recovery and Navigation}
    
\author{
\IEEEauthorblockN{
Vignesh Rajagopal${}^{1}$ and Rohan Chandra${}^{1}$\\[0.2em]
Kasun Weerakoon${}^{2}$, Gershom Seneviratne${}^{2}$, Pon Aswin Sankaralingam${}^{2}$\\
Mohamed Elnoor${}^{2}$, Jing Liang${}^{2}$, and Dinesh Manocha${}^{2}$
}
\thanks{${}^{1}$ Authors are with the Dept. of Computer Science at the University of Virginia. Contact email: {\tt\footnotesize atk9sb@virginia.edu}}
\thanks{${}^{2}$ Authors are with the Dept. of Computer Science at the University of Maryland College Park.}
}

\makeatletter
\let\NAT@parse\undefined
\makeatother
\usepackage[dvipsnames]{xcolor}
\usepackage[colorlink
s=true,
    citecolor=orange,
    linkcolor=red,
    urlcolor=blue]{hyperref}

\begin{document}
\raggedbottom

\maketitle
\thispagestyle{empty}
\pagestyle{empty}

\begin{abstract}

We present DR. Nav (Dead-End Recovery-aware Navigation), a novel approach to autonomous navigation in scenarios where dead-end detection and recovery are critical, particularly in unstructured environments where robots must handle corners, vegetation occlusions, and blocked junctions. DR. Nav introduces a proactive strategy for navigation in unmapped environments without prior assumptions. Our method unifies dead-end prediction and recovery by generating a single, continuous, real-time semantic cost map. Specifically, DR. Nav leverages cross-modal RGB–LiDAR fusion with attention-based filtering to estimate per-cell dead-end likelihoods and recovery points, which are continuously updated through Bayesian inference to enhance robustness. Unlike prior mapping methods that only encode traversability, DR. Nav explicitly incorporates recovery-aware risk into the navigation cost map, enabling robots to anticipate unsafe regions and plan safer alternative trajectories. We evaluate DR. Nav across multiple dense indoor and outdoor scenarios and demonstrate an increase of 83.33\% in accuracy in detection, a 52.4\% reduction in time-to-goal (path efficiency), compared to state-of-the-art planners such as DWA, MPPI, and Nav2 DWB. Furthermore, the dead-end classifier functions 
\end{abstract}

\section{Introduction}  \label{sec:Intro}

Robot navigation has been applied in diverse domains such as logistics~\cite{alverhed2024autonomous, chen2021adoption}, services~\cite{berezina2019robots, camarillo2004robotic, rivero2023robotic}, and exploration in hazardous environments~\cite{trevelyan2016robotics, petereit2019robdekon, kalita2020exploration}. Although robot navigation has been studied for decades~\cite{Fox1997DWA, ofvo, densecavoid, graspe, adaptiveon}, mapless navigation in unstructured environments remains a significant challenge~\cite{dtg,vl-tgs,behav, mtg,seneviratne2024cross, chandra2025deadlock}. In particular, the problem of detecting and navigating around dead-ends in unexplored environments spaces has received limited attention~\cite{wang2008fuzzy, kang2011genetic}. Robots must often decide whether an unseen corridor or trail will remain open or terminate in a dead-end, a capability that is crucial for navigation as robots can easily become trapped~\cite{wang2008fuzzy, kang2011genetic, densecavoid, steinmetz2017search}, and recovery can typically be both time-consuming and challenging~\cite{steinmetz2017search,kang2011genetic} as robots must make predictions based on partial observations while simultaneously planning safe and efficient paths~\cite{chandra2023socialmapf, raj2024rethinking, zinage2025decentralized}.

\begin{figure}[t]
    \centering
    \includegraphics[width=1.0\columnwidth]{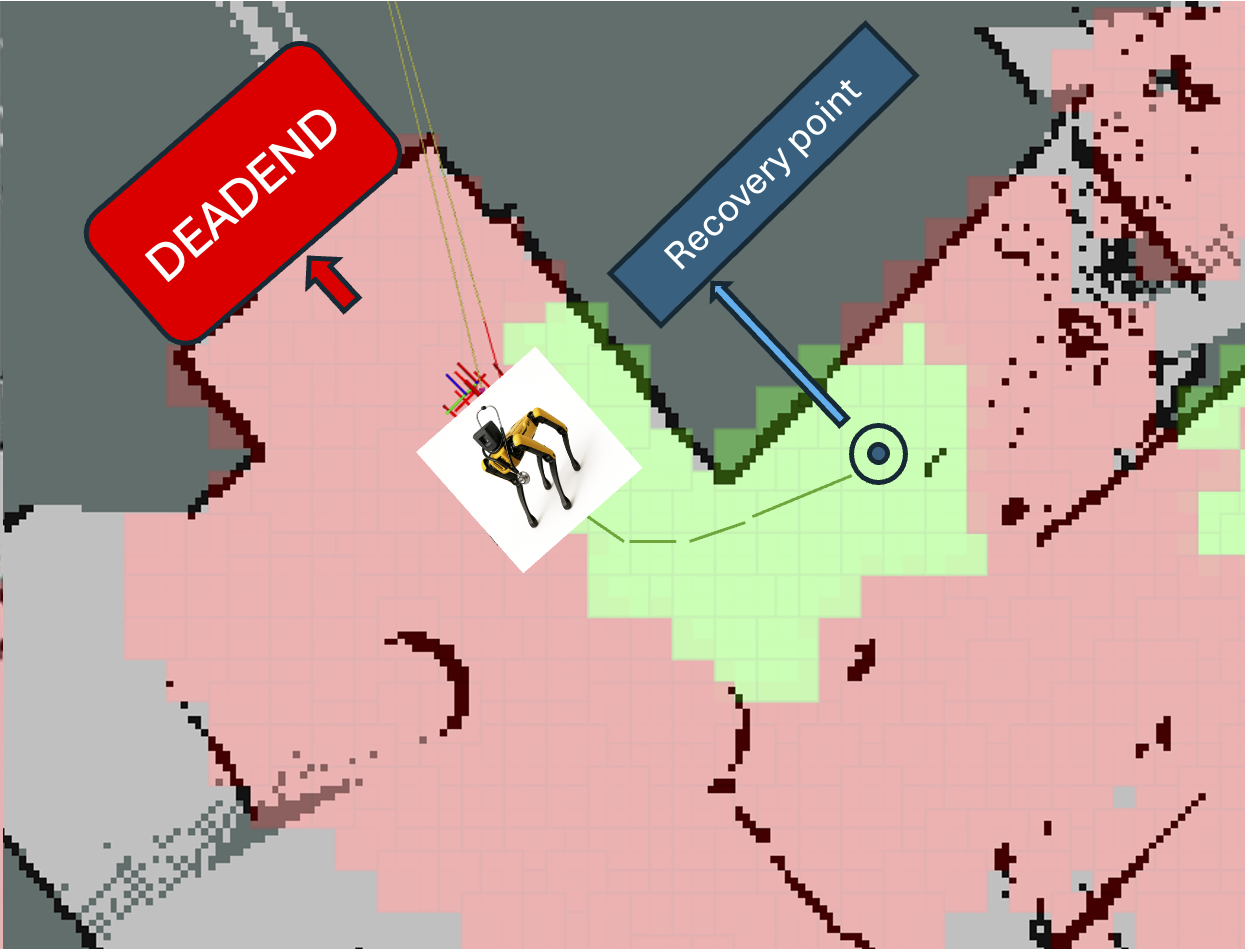}
    \caption{\small{The continuous cost map encodes environment semantics at the cell level, where green regions represent navigable areas and red regions correspond to predicted dead-ends or risky zones and the recovery point as blue cylinders . Unlike a binary occupancy grid, this representation continuously updates as new RGB–LiDAR observations arrive, allowing the robot to reason about both safety and recoverability. In this example, the robot (yellow marker) progresses forward as long as at least one side of its camera view remains open. When the surrounding views become fully blocked, the system detects the impending dead-end (red region ahead) and proactively backtracks to a pre-identified recovery point (blue marker). This illustrates how the continuous cost map not only predicts unsafe regions in advance but also guides the robot toward safe alternatives in real time.).
    }}
    \label{fig:cover}
\end{figure}


Indoor scenarios such as blind corner corridors and blocked T-junctions~\cite{wang2008fuzzy}, as well as outdoor scenarios involving dense vegetation or terrain occlusions that often lead to dead-ends~\cite{graspe, terrapn, chandra2025multi}, pose a risk of robots becoming trapped~\cite{chandra2025deadlock, chandra2025multi, gouru2024livenet, mahadevan2025gamechat, chen2025livepoint, francis2025principles}. Entering such regions forces risky U-turns or lengthy backtracking, which wastes time and increases the likelihood of failure. Predicting these situations~\cite{wang2008fuzzy, steinmetz2017search, chandra2024multi} in advance is difficult due to limited sensing range, partial observability, and ambiguous visual appearances. Moreover, existing local planners~\cite{crowdsteer, ofvo, Fox1997DWA} typically optimize only short-horizon feasibility and collision avoidance, without anticipating dead-ends or reasoning about recovery strategies.

Most existing navigation pipelines are reactive~\cite{steinmetz2017search, kang2011genetic}: robots attempt recovery only after becoming trapped. Proactive navigation that avoids entering dead-ends requires analyzing scene risks in advance. Current risk-aware navigation~\cite{cai2022riskawareoffroadnavigationlearned,10606099,Randriamiarintsoa_2023} has made progress by learning terrain traversability or modeling uncertainty to penalize risky motions, but these approaches~\cite{cai2022riskawareoffroadnavigationlearned, 10606099} focus only on terrain safety and uncertainty, without explicitly modeling dead-ends or recovery strategies. Existing recovery methods~\cite{steinmetz2017search, cao2024low} also fail to evaluate which candidate strategy is most viable in terms of occlusion, proximity to the goal, or energy efficiency. Thus, while prior work has advanced risk-aware navigation and semantic reasoning, no existing framework provides a unified solution for proactively predicting dead-ends and selecting recovery points in real time across both indoor and outdoor environments.

\noindent {\bf Main Contributions:} In this paper, we present Semantic-Geometric Representations for Proactive Dead-End Recovery and Navigation (DR. Nav), a framework that shifts navigation from reactive escape to proactive avoidance. DR. Nav fuses RGB and LiDAR observations through cross-modal attention to infer per-cell dead-end probabilities in a local semantic map. This map is continuously updated with Bayesian filtering, ensuring stable predictions under occlusion and noise. Beyond traversability, it explicitly encodes recovery points and integrates them into the planning loop. A short-horizon goal generator scores candidate headings using both geometric feasibility and an Expected Dead-End Exposure (EDE) term, enabling proactive re-routing before the robot commits to blocked paths. Some novel aspects of our work include:

\begin{itemize}
    \item A \textbf{proactive navigation framework} that enables robots to predict dead-ends and plan recoveries in cul-de-sacs, blind corners, vegetation occlusions, and so on. Current state-of-the-art is reactive, and only recover from dead-ends after getting trapped. 

    \item A novel \textbf{representation learning} algorithm that consists of an RBG (semantic) - lidar (geometric) fusion using cross-attention.
    
    \item A novel \textbf{cost map} that continuously encodes per-cell dead-end probabilities and recovery points learned via Bayesian log-odds filtering for long-horizon navigation.
    
    \item A \textbf{goal generation strategy} that minimizes expected dead-end exposure while continuously selecting safe recovery paths, improving both success rates and navigation path efficiency compared to reactive baselines.

\end{itemize}

\section{Related Works}
In this section, we review prior literature on dead-end detection and recovery, risk-aware costmaps, semantic–geometric fusion, and multimodal cross-attention. We also position our approach in relation to these works, highlighting how it extends and complements them.

\subsection{Dead-End Detection and Recovery}
Early navigation methods such as the Dynamic Window Approach (DWA)~\cite{Fox1997DWA} and the Global Dynamic Window~\cite{Brock1999GDWA} provided robust short-horizon collision avoidance, but they often fell into local minima or dead-ends due to limited foresight. Classical occupancy grid planners~\cite{Moravec1985OG,Thrun2005ProbRobotics} likewise relied on reactive recovery behaviors when the robot became trapped.  
Recent advances attempt to proactively anticipate failures. Cai \emph{et al.}~\cite{Cai2022RiskAwareSpeedMap} learned terrain speed distributions to predict unsafe zones, while EVORA~\cite{Cai2023EVORA} applied evidential deep learning to quantify traversability uncertainty. Ji \emph{et al.}~\cite{Ji2022PAAD} introduced PAAD, a proactive anomaly detector using multi-sensor fusion, and Schreiber \emph{et al.} (ROAR) extended this toward occlusion handling. Geometric approaches compute convex safe corridors using IRIS~\cite{Deits2014IRIS}, improved by faster SDP relaxations~\cite{Werner2024FasterIRIS} and nonlinear optimization~\cite{Petersen2023GrowingConvex}, while Liu \emph{et al.}~\cite{Liu2017SFC} designed safe-flight corridors for aerial robots.  

\noindent\textbf{How DR. Nav advances the state-of-the-art:} Our approach differs by predicting per-cell \emph{dead-end probabilities} directly in the costmap, independent of a specific planned trajectory. This planner-agnostic view enables earlier avoidance than reactive heuristics and greater flexibility than plan-conditioned risk predictors.

\subsection{Risk-Aware Costmaps and Bayesian Filtering}
Occupancy grid mapping with Bayesian log-odds updates~\cite{Moravec1985OG,Thrun2003ForwardSensorOG} remains foundational, offering stable probabilistic representations under sensor noise. However, standard occupancy maps classify cells as free or occupied without encoding traversal \emph{risk}. Laconte \emph{et al.} proposed the Lambda Field~\cite{Laconte2020LambdaField}, extending occupancy maps to estimate continuous collision risk.  

\noindent\textbf{How DR. Nav advances the state-of-the-art:} Our continuous costmap generalizes this idea: instead of collision likelihood, we maintain Bayesian-filtered estimates of dead-end probability. By using log-odds fusion, the system integrates multi-sensor evidence over time and stabilizes predictions under occlusion. Unlike static occupancy grids, our costmap is dynamic and explicitly encodes recovery points, allowing planners to minimize expected dead-end exposure.

\subsection{Semantic–Geometric Fusion}
LiDAR provides geometric precision, but vision supplies rich semantics. Recent multimodal works demonstrate the advantages of combining them. Buckeridge \emph{et al.} used semantic segmentation fused with LiDAR for sidewalk navigation, while Urrea \emph{et al.} augmented costmaps with class-dependent inflation using RGB–LiDAR fusion. In autonomous driving, BEVFusion~\cite{Liang2022BEVFusion} and TransFusion~\cite{Bai2022TransFusion} unify LiDAR and camera features for robust 3D perception.  

\noindent\textbf{How DR. Nav advances the state-of-the-art:} Our system leverages \emph{multi-camera + LiDAR} fusion specifically for dead-end awareness: semantic cues from cameras (e.g., doorways, vegetation) are aligned with LiDAR geometry to assess trap likelihood. This fusion enables detection of subtle cues (like occluded cul-de-sacs) that either modality alone would miss.

\begin{figure*}[t]
    \centering
    \includegraphics[width=2\columnwidth]{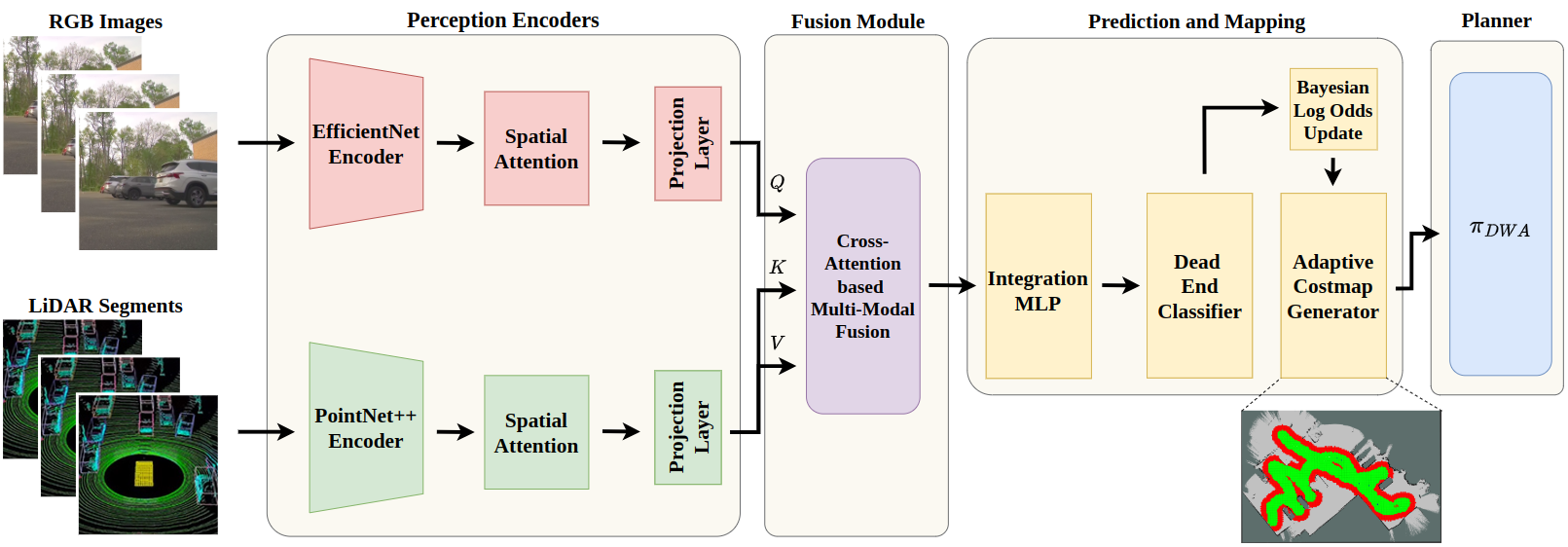}
    \caption{\small{Our method uses a multi-modal sensor fusion architecture for  costmap computations and  autonomous navigation. Our proposed model processes RGB images through an EfficientNet encoder and LiDAR point clouds through a PointNet encoder, applying dual-stage spatial attention mechanisms to enhance feature representations. We perform cross-modal fusion to integrate the enhanced multi-modal features, which are then processed through an integration MLP to generate path classification, dead-end detection, and cost map outputs for autonomous navigation tasks.  
    }}
    \label{fig:cover}
\end{figure*}

\subsection{Cross-Attention for Multimodal Fusion}
Transformer-based cross-attention has proven powerful for aligning heterogeneous modalities. TransFuser~\cite{Prakash2021TransFuser,Chitta2023TransFuserPAMI} applied multi-head attention to fuse image and LiDAR features for end-to-end driving, while BEVFusion~\cite{Liang2022BEVFusion} and TransFusion~\cite{Bai2022TransFusion} achieved strong 3D detection by attending across modalities. Zhao \emph{et al.}~\cite{Zhao2022CrossModalLoc360} further showed cross-attention improves pose estimation by linking 360° images with point clouds.  

\noindent\textbf{How DR. Nav advances the state-of-the-art:} We adopt multi-head cross-attention for recoverability awareness: RGB attends to LiDAR geometry and vice versa at the costmap cell level, yielding stable, planner-agnostic estimates of both dead-end risk and recovery opportunities. This differs from prior detection- or localization-focused fusion by directly shaping a navigation costmap.

\subsection{Planners and Recovery Behaviors}
Local planners such as DWA~\cite{Fox1997DWA} and MPPI~\cite{Williams2017ITMPC} optimize short-horizon feasibility of actions in the constrained environment but lack explicit cul-de-sac anticipation. Recovery modules in ROS navigation~\cite{Zheng2016NavGuide} typically rotate-in-place or clear costmaps reactively once the robot is stuck.  
The Dynamic Window Approach (DWA)~\cite{Fox1997DWA} evaluates feasible velocity commands in real time to ensure collision-free motion, while its extensions~\cite{Brock1999GDWA} improve trajectory smoothness and dynamic feasibility. Similarly, sampling-based methods such as Model Predictive Path Integral control (MPPI)~\cite{Williams2017ITMPC} optimize short-horizon costs over randomized trajectories, enabling agile maneuvers even in clutter. Despite their success, these planners inherently operate on local geometric cues and short prediction horizons. As a result, they cannot anticipate long-range traps such as cul-de-sacs or narrow corridors hidden by occlusions, leading to risky late-stage reversals or oscillations. Recovery behaviors in standard frameworks (e.g., ROS navigation~\cite{Zheng2016NavGuide}) address failures only after entrapment, by rotating in place or clearing costmaps.  

\noindent\textbf{How DR. Nav advances the state-of-the-art:} By contrast, our goal generator biases planning toward routes with low expected dead-end exposure and preemptively triggers recovery before failure, complementing existing planners rather than replacing them. Our method annotates Bayesian-filtered dead-end probabilities and recovery points into a continous costmap; we provide these reactive planners with long-range semantic awareness. The goal generator then biases trajectories away from high-risk corridors and preemptively triggers recovery before commitment to a dead-end. In doing so, our framework transforms reactive planners into proactive navigators capable of safe and efficient operation in both urban and vegetated environments.

\bibliographystyle{IEEEtran}

\section{Problem Formulation}


We represent our problem via the tuple $\left \langle \mathcal{M}, \mathcal{U}, \mathcal{J}, \mathcal{O} \right \rangle$. Here, $\mathcal{M}$ is a local map created on-the-fly using lidar-based SLAM, that is, a global map is not assumed. $\mathcal{M}$ is spatially discretized into square cells $c \in \mathcal{M}$ of size $0.01m^2$ which can be traversable, occupied, or uncertain due to occlusion. Our approach works with standard off-the-shelf planners (MPC, DWA, etc) and therefore our action space $\mathcal{U}$ consists of linear and angular velocity, $v$ and $\omega$, respectively. Over a time horizon $1:t$, the trajectory, $\tau$ generated is evaluated by a cost function $\mathcal{J}(\tau)$. $\mathcal{O}_t$ represents a time-synchronized RGB image and a lidar pointcloud scan. 


\subsection{Objective}

At every $t$, given $\mathcal{O}_t$, our goal is to predict a costmap $\mathcal{C}_t$ that annotates every $c\in\mathcal{M}$ with the likelihood of ending up in a dead-end. This probability is denoted as $P_{\text{dead}}(c \mid \mathcal{O}_{1:t})$. We then solve the following optimization problem:

\begin{equation}
\arg\min_{\substack{v_{t:t+H} \\ \omega_{t:t+H}}} 
    J_{\text{geom}}(\tau) 
    + \lambda \cdot \mathbb{E}_{c \in \tau}\big[P_{\text{dead}}(c \mid \mathcal{O}_{1:t})\big],
    \label{eq: cost_func1}
\end{equation}

where $H$ corresponds to a fixed horizon and $J_{\text{geom}}(\tau)$ encodes standard geometric constraints (collision avoidance, smoothness, and kinematic feasibility)
The scalar $\lambda \geq 0$ controls the influence of semantic risk relative to geometric feasibility. We integrate over $v_{t:t+H}, \omega_{t:t+H}$ to generate local goals in $\mathcal{M}$ that minimize expected dead-end exposure while proactively triggering recoveries before the robot commits to unsafe paths.

\section{Our Approach}
\label{sec: approach}

At each time $t$, the robot receives partial observations from onboard multi-camera and LiDAR sensors, denoted as $\mathcal{O}_t$, at time $t$. These observations are first individually encoded into mixed semantic-geometric feature representations via transformer-based neural networks (Section~\ref{subsec: tokens}), followed by fusing these features via cross-attention (Section~\ref{subsec: fusion}) to generate new fusion features. The fusion features are used to train a classifier using Bayesian log-odds filtering (Section~\ref{subsec: costmap}) that predicts, in real time, the probability of a cell leading to a dead-end, denoted as $P_{\text{dead}}(c \mid \mathcal{O}_{1:t})$, while simultaneously identifying potential recovery points that enable safe escape strategies. We then embed $P_{\text{dead}}(c \mid \mathcal{O}_{1:t})$ in a cost function and use classical off-the-shelf planners to generate short-term local goals that proactively avoid the dead-end inducing cells in realtime (Section~\ref{subsec: planning}).


\subsection{From Perception ($\mathcal{O}_{1:t}$) to Semantic-Geometric Features}
\label{subsec: tokens}
Each sensing modality is transformed into a compact \emph{token}, defined as a fixed-length feature vector rather than an image or 
heatmap. For the camera streams, the input at each frame is an RGB image 
$I \in \mathbb{R}^{H \times W \times 3}$. A lightweight convolutional backbone extracts patch-level descriptors $\{f_k\}$ together with a global descriptor $g$. An attention mechanism then computes weights 
$\alpha_k = \mathrm{softmax}(g^\top W f_k)$, which quantify the contribution of each patch to navigation. The final camera token is 
$z_{\text{img}} = \sum_k \alpha_k f_k \in \mathbb{R}^{d}$, produced once per camera view at the native camera frame rate. While the weights $\alpha_k$ can be visualized as a saliency heatmap for interpretability, the model only propagates the fused feature vector $z_{\text{img}}$ to downstream modules. On the LiDAR side, the input at each scan is a set of points $P = \{p_i \in \mathbb{R}^3\}_{i=1..N}$. A streamlined PointNet-style encoder maps each point to a local feature $h_i$ and aggregates them into a global context $u$. A learned attention mechanism assigns scores 
$\beta_i = \mathrm{softmax}(\phi([h_i;u]))$ to emphasize geometrically stable returns while suppressing noisy measurements. The LiDAR token is then computed as $z_{\text{lidar}} = \sum_i \beta_i h_i \in \mathbb{R}^{d}$, produced once per LiDAR scan at the sensor’s operating frequency. These tokens, one per camera view and one per LiDAR scan are time-synchronized and form the inputs to the cross-modal fusion module.

\subsection{Cross-Modal Fusion of $z_{\text{lidar}}$ and $z_{\text{img}}$ }
\label{subsec: fusion}
The fusion module aligns and conditions RGB and LiDAR information in real time. At its heart lies a bi-directional cross-attention mechanism: image tokens query LiDAR features to highlight traversable structures, while LiDAR tokens query image features to disambiguate occluded or 
texturally ambiguous regions. Residual connections and normalization ensure stability and low-latency operation. Formally, let $z_{img} \in \mathbb{R}^{N \times d}$ denote the set of $N$ visual tokens extracted from multi-camera RGB images, and $z_{lidar} \in \mathbb{R}^{M \times d}$ the $M$ tokens derived from the LiDAR point cloud encoder. 
Cross-attention is performed in both directions:

\begin{equation}
\begin{aligned}
    z_{img} &= \\
    &\text{Attn}(Q = z_{img}W_q,\, K = z_{lidar}W_k,\, V = z_{lidar}W_v), \\
    z_{lidar} &= \\
    &\text{Attn}(Q = z_{lidar}W_q',\, K = z_{img}W_k',\, V = z_{img}W_v').
\end{aligned}
\end{equation}

where  $W_q, W_k, W_v$ are learned projections. The two updated streams are then merged through a residual feed-forward block($FFN$) :
\begin{align}
    z_{fuse} = \text{FFN}\left([\hat{z}_{img} \, \Vert \, \hat{z}_{lidar}]\right),
\end{align}
yielding a joint embedding that encodes both semantic cues from vision and geometric structure from Lidar.

The fused representation from multiple cameras is integrated through a lightweight feed-forward block, yielding a joint embedding that carries both semantic cues (from vision) and geometric grounding (from LiDAR). This embedding directly feeds into the navigation heads. 

\subsection{Continuous Semantic Cost Map ($\mathcal{C}_t$)}
\label{subsec: costmap}

The fused tokens, $z_{fuse}$, are projected into a grid-based semantic cost map in the robot’s frame. 
Each grid cell $c$ is associated with a latent probability $p_{dead}(c)$ of being a dead-end at time $t$. To ensure temporal consistency, we adopt a Bayesian log-odds update rather than directly averaging probabilities. 

\vspace{7pt}

\noindent\textbf{Bayesian Log-Odds Filtering:} Let $l_t(c)$ denote the log-odds representation of cell $c$:
\begin{align}
    l_t(c) &= \log \frac{p_{dead}(c)}{1 - p_{dead}(c)}.
\end{align}
Given a new observation with predicted probability $\hat{p}_{dead}(c)$ from the fusion module, 
The recursive update is
\begin{align}
    l_t(c) &= l_{t-1}(c) + \log \frac{\hat{p}_{dead}(c)}{1 - \hat{p}_{dead}(c)} - l_0,
\end{align}
where $l_0 = \log \frac{p_0}{1-p_0}$ encodes the prior belief. 
The posterior probability is then recovered as
\begin{align}
    p_{dead}(c) &= \frac{1}{1 + \exp\big(-l_t(c)\big)}.
\end{align}

This formulation offers two key advantages: $(i)$ stability under noise and occlusion, since log-odds 
accumulate evidence over time, and $(ii)$ fast adaptation to new observations, as contradictory 
evidence rapidly shifts the posterior away from outdated beliefs. 

In practice, updates are performed at 10~Hz, producing a probability field over the environment. 
The resulting cost map can be visualized as a dynamic heatmap, with green cells denoting safe passage 
and red cells indicating likely dead-ends. Unlike static traversability maps, this representation evolves 
continuously with perception, encoding both risk and recovery opportunities. 
This dynamic Bayesian filtering constitutes a core contribution of our framework, enabling robots to 
anticipate unsafe regions while maintaining robust navigation performance in highly uncertain environments.

\vspace{7pt}

\noindent\textbf{Dead-End Posterior Per Cell ($P_{dead}(c,t)$):}
The posterior probability can say that given a cell c is a dead-end/open at time t.
Let $\hat{p}_{dead}(c)\in(0,1)$ be the fusion module's instantaneous estimate and
$l_t(c)$ the log-odds state. We update
\begin{align*}
& l_t(c) = \mathrm{clip}\!\left(l_{t-1}(c) + \log\frac{\hat{p}_{dead}(c)}{1-\hat{p}_{dead}(c)} - l_0, l_{\min}, l_{\max}\right),\\
&P_{\text{dead}}(c,t) = \sigma\!\big(l_t(c)\big) = \frac{1}{1+\exp(-l_t(c))}.
\end{align*}
where $\sigma$ is the logistic sigmoid function, which maps the log odds into probability


\vspace{7pt}

\noindent\textbf{Expected Dead-End Exposure (EDE):} The Expected dead-end exposure for a candidate path $\gamma$ aggregates those per-cell probabilities along the short horizon.

For a candidate control  $\theta=(v,\omega)$ inducing a short-horizon path $\gamma_\theta=\{c_k\}_{k=1}^{K}$ in the grid, we define the additive exposure
\begin{align*}
\mathrm{EDE}(\theta) = \sum_{k=1}^{K} w_k\, P_{\text{dead}}^{\text{foot}}(c_k, t)\
\end{align*}
where $w_k=\exp(-\lambda s_k)$ discounts distant cells. In continuous form with state $x(s)$ along $\gamma_\theta$,
\begin{align*}
\mathrm{EDE}(\theta) = \int_{0}^{H} w(s)\, P_{\text{dead}}^{\text{foot}}(x(s), t)\, ds.
\end{align*}

\subsection{Planning with EDE}
\label{subsec: planning}
The cost map is consumed by a short-horizon goal generator. 
For each candidate waypoint within a 3–5\,m horizon, the planner evaluates a joint score:
\begin{equation}
     J_{\text{geom}}(\theta) + \lambda \cdot EDE(\theta),
    \label{eq: cost_function}
\end{equation}
where $J_{\text{geom}}$ accounts for geometric feasibility (collision, smoothness, range bias), 
and $EDE$ 
integrates the predicted dead-end probabilities along the trajectory. 
The weight $\lambda$ controls the influence of semantic risk. 
Instead of directly executing $\theta^\star$, we roll out all admissible controls 
$\theta \in \Theta_t^{\text{adm}}$ (\textit{adm is the admissible action space}) for horizon $H$, and record their endpoints 
$g(\theta) = x_K \in \mathbb{R}^2$. Each endpoint represents a candidate waypoint. The score is defined at the control level and then associated with its endpoint: 
\begin{align}
\mathrm{Score}(g(\theta)) &= J_{\text{geom}}(\theta) + \lambda\,\mathrm{EDE}(\theta)
\end{align}
The short-horizon goal is then chosen as
\begin{align}
g^\star &= \arg\min_{g(\theta)} \;\mathrm{Score}(g(\theta))
\end{align}
The score is computed over the generating control
$\theta$, but is attached to its endpoint $g(\theta)$.
\noindent The DWA controller samples
admissible controls to track $g^\star$ while still enforcing collision-free rollouts against the cost map. In this way, the semantic cost map influences navigation 
twice: first through the goal generator (which avoids dead-end exposure), and 
then through DWA (which ensures safe geometric tracking of the chosen waypoint). For purely reactive baselines, we set $\lambda=0$ and $\mathrm{EDE}$ is ignored, recovering the standard DWA objective.
If all candidate rays are blocked or high-risk, the planner switches to semantic recovery, 
using recovery points inferred from the cost map. 
This tight perception–planning loop enables the robot to choose paths that minimize 
expected dead-end exposure, anticipating failures rather than reacting to them. By compressing RGB and LiDAR into attention-driven tokens, fusing them bi-directionally in real time, 
updating a Bayesian semantic cost map, and integrating this into an EDE-aware planner, 
our approach achieves early dead-end detection, fewer stalls, and more energy-efficient navigation 
compared with purely reactive baselines. 

\section{Experiments}

In this section, we investigate the following questions: 

\begin{enumerate}
    \item Can DR. Nav detect potential dead-end scenarios much earlier and more reliably than state-of-the-art planners.
    \item Can DR. Nav result in more efficient trajectories as a result of proactive dead-end detection and recovery?
\end{enumerate}

\subsection{Scenarios, Baselines, and Metrics}
\label{sec:baselines}
We demonstrate the benefits of our approach in dead-end detection, risk estimation, and recovery navigation, which helps robots escape dead-end traps. We conduct experiments on ROSBag replays collected on a Ghost robot in various indoor and outdoor environments. These environments consist of naturally occurring dead-ends, such as elevators, loading docks, walls, and parked cars, that represent U-shaped cul-de-sacs, T-junctions, and straight corners. 
Navigation quality is quantified using three metrics:
\begin{itemize}
    \item {Distance (m):} The total distance traveled by the robot from start to end. Shorter paths indicate more efficient planning.
    \item {Path efficiency:} Defined as the ratio of the straight-line distance to the actual path length. 
    Higher efficiency implies fewer detours and unnecessary maneuvers.
    \item Avg. speed (m/s): The mean forward velocity during successful runs captures how often the robot stalls or hesitates due to unsafe regions.
\end{itemize}
Detection quality is quantified using three metrics:
\begin{itemize}
    \item {Accuracy (\%):} is the percentage of correctly classified tiles 
(OPEN vs.\ BLOCKED) over $N$ labeled samples. 
    \item {Latency (s/img):} measures the average wall-clock time required to 
process one image, reported in seconds per image (s/img).
    \item {Throughput ($Hz$):} is the reciprocal of latency, expressed in 
frames per second (Hz), where higher values indicate faster inference. 
\end{itemize}
We compare our approach with four planners that all share the same short-horizon waypoint sampler and SLAM/local costmap; only the semantic weight $\lambda$ and the controller differ:

\begin{itemize}
    \item DWA + DR. Nav cost (ours), which minimizes a geometric term plus Expected Dead-end Exposure (EDE).
    \item DWA (vanilla) with $\lambda=0$.
    \item DWB (Nav2 default) with stock critics.
    \item MPPI (Nav2) without semantics.
\end{itemize}

For semantics, we compare our RGB–LiDAR fusion (DR. Nav) against image-only large vision-language models (VLMs) used as reference detectors: ChatGPT, Gemini, and CLIP (zero-shot text–image similarity for ``dead-end ahead'' vs ``clear path''). VLMs are not used for control, rather, they serve to benchmark per-frame dead-end detection.

\subsection{Role of DR. Nav Towards Improving Navigation Efficiency}

\begin{figure}[ht]
    \centering
    \includegraphics[width=1.0\linewidth]{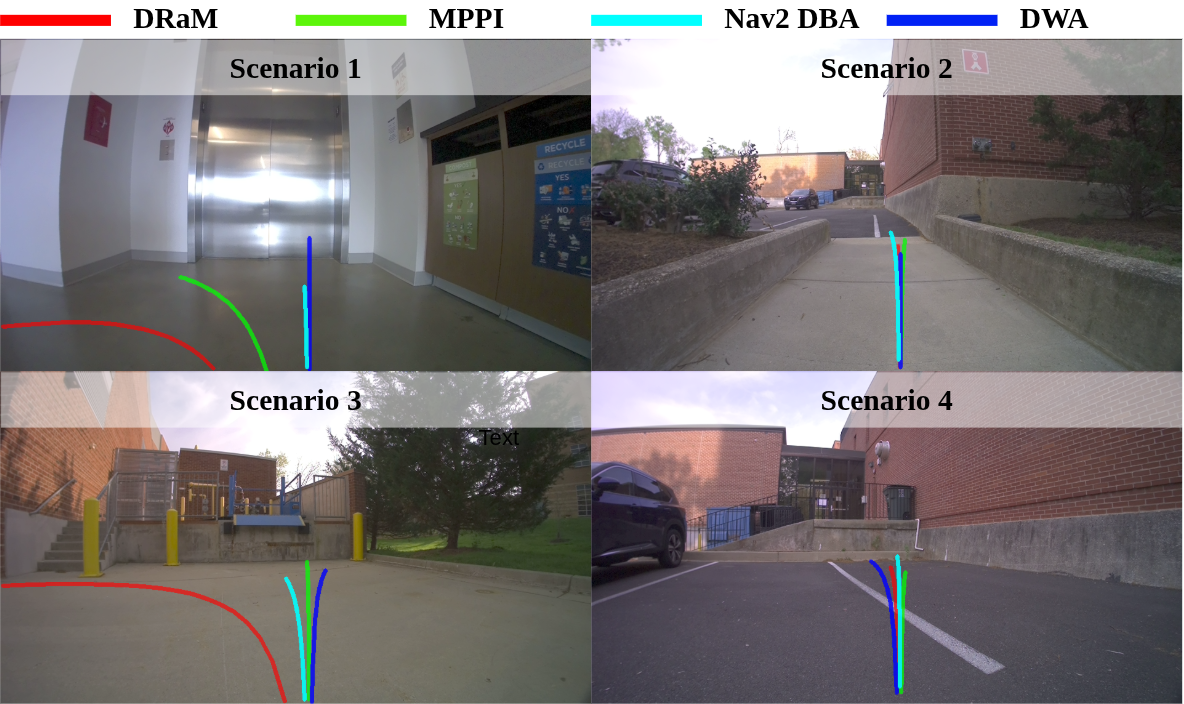}
    \caption{\small{Qualitative results. These qualitative results show that DR. Nav introduces proactive divergent trajectories compared to current planners that fail to account for the dead-ends, become trapped, and then backtrack.}}
    \label{fig:qualitative_detection}
\end{figure}

As shown in Table~\ref{tab:quantitative_navigation}, the dist(m) metric shows that given an input scenario from a rosbag, DR. Nav+DWA yields the shortest, most direct paths and reduces wandering near predicted dead-ends
, indicating a risk-aware, efficiency-first behavior that we will combine with improved recovery to improve the goal completion in a shorter and energy efficient way.
As shown in Figure~\ref{fig:qualitative_detection}, it illustrates four representative navigation scenarios comparing the trajectories produced by different planners: our proposed DR. Nav (\textcolor{red}{\textbf{red}}), MPPI (\textcolor{green}{\textbf{green}}), Nav2 DBA (\textcolor{cyan}{\textbf{cyan}}), and the standard DWA (\textcolor{blue}{\textbf{blue}}). Each sub-figure shows the robot’s camera view overlaid with the predicted path, highlighting how different methods respond to dead-ends or constrained passages. We observe that in various dead-end scenarios, DR. Nav is able to plan trajectories that avoid the dead-end well in advance.





\begin{figure}[t]
    \centering
    \includegraphics[width=1.0\columnwidth]{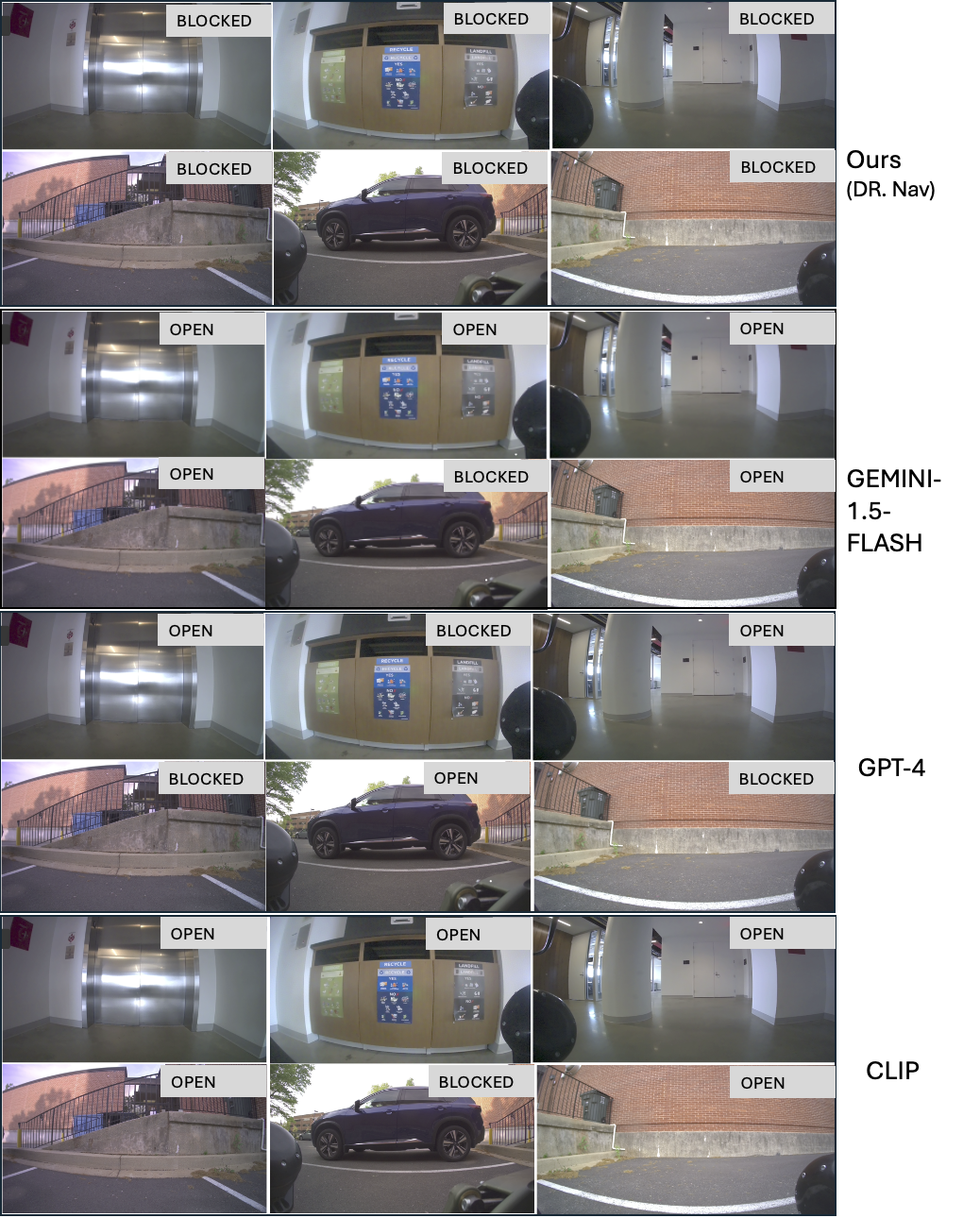}
    \caption{\small{Qualitative results of dead-end detection and risk estimation}}
    \label{fig:qualitative_navigation}
\end{figure}

\begin{table}[t]
\centering
\resizebox{\columnwidth}{!}{%
\begin{tabular}{rccc} 
\toprule
\textbf{Model}            & {\textbf{Accuracy} (\%) $\uparrow$} & {\textbf{Latency} (s/img) $\downarrow$} & {\textbf{Throughput} (Hz) $\uparrow$} \\
\midrule
\textbf{DR. Nav head} & \textbf{{83.33}} & \textbf{0.033} & \textbf{30.0} \\
GPT                        & {66.66} & 2.000 & 0.50 \\
Gemini                     & {33.33} & 2.500 & 0.40 \\
CLIP                       & {33.33} & 2.440 & 0.41 \\
\bottomrule
\end{tabular}%
}
\caption{Dead-end detection on exemplars from Fig.~\ref{fig:qualitative_navigation}. 
Accuracy is the fraction of correct \texttt{OPEN}/\texttt{BLOCKED} decisions over $N$ labeled tiles. Latency is per-image wall time; Hz is throughput (higher is better).}
\label{tab:detection_metrics}
\end{table}


\begin{table}[t]
\centering
\scriptsize

\begin{tabular}{rccc} 
\toprule
\textbf{Method}  & {\textbf{Dist.}~(m)~$\downarrow$}& {\textbf{Speed}~(m/s)~$\uparrow$} & {\textbf{Eff.}~$\uparrow$} \\
\midrule
DR. Nav + DWA   & \bfseries \textbf{559.5}  & 0.157 & \bfseries \textbf{0.524} \\
MPPI         & 1966.1 & \bfseries 0.240 & 0.294 \\
Nav2\_DWB    & 2325.1 & \bfseries 0.354 & 0.114 \\
Vanilla DWA  & 2725.5           & 0.000 & 0.000 \\
\bottomrule
\end{tabular}%
\caption{Quantitative results over 5 bag replay runs (goal tolerance $0.75$\,m). 
Arrows indicate a better direction. DR. Nav\,+\, DWA yields the shortest path.}
\label{tab:quantitative_navigation}
\end{table}

\subsection{DR. Nav's Dead-end Detection and Risk Estimation}

Figure~\ref{fig:qualitative_navigation} presents a qualitative comparison of dead-end detection across different vision language models (Ours, GEMINI-1.5-FLASH, GPT-4 \& CLIP). Each row shows predictions for diverse indoor and outdoor environments, with labels indicating whether the path ahead is classified as \texttt{OPEN} or \texttt{BLOCKED}.

DR. Nav consistently identifies blocked regions with higher reliability. For example, correctly flagging elevators, closed corridors, and obstructed outdoor ramps as BLOCKED. Existing methods, however, frequently misclassify these scenarios as OPEN, which would result in unsafe navigation behaviors if used in real-time planning. Notably, GEMINI-1.5-flash and GPT-4 occasionally detect blocked cases but suffer from inconsistencies, while CLIP often fails to capture the true navigational risk, labeling most environments as OPEN. This comparison highlights that our approach not only improves prediction accuracy but also maintains stability across diverse contexts, ensuring that the robot proactively avoids dead-ends before committing to unsafe trajectories.

\subsection{Analysis and Discussion}
Fig.~\ref{fig:qualitative_navigation} presents four representative scenarios that we use to qualitatively compare DR. Nav with the baselines described in Section~\ref{sec:baselines}. In Scenario 1, DR. Nav turns away from the dead-end near the elevator while every baseline method keeps going into the dead-end. In Scenario 2, all methods avoid the walls while heading along the path. We use this to show that normal behavior too can be observed without just turning away from dead-ends. In Scenario 3, DR. Nav turns away from the dead-end leading to the loading deck that the robot would get stuck in while all other methods go into the dead-end. In Scenario 4, this is a case where the dead-end was not detected and all methods go into the dead-end. 
 
Furthermore, DR. Nav{+}DWA produces by far the shortest paths, as shown in the Table~\ref{tab:quantitative_navigation}, indicating more direct, low-cost motion, and path efficiency is lower compared to the other classical planners. DR. Nav+DWA penalizes wandering/backtracking, effectively achieving higher efficiency. 

Furthermore, the RGB--LiDAR fusion model is further benchmarked against vision-language models that operate purely on images, including GPT-4, Gemini-1.5-flash, and CLIP. Table~\ref{tab:detection_metrics} reports dead-end detection accuracy, inference latency, and throughput across our RGB–LiDAR fusion model and image-only vision–language models. DR. Nav head achieves the highest accuracy (83.3\%), substantially outperforming GPT (66.7\%) and Gemini/CLIP (33.3\%). This indicates that fusing geometric and semantic cues enables more reliable early detection of blocked passages than relying on images alone. In addition to accuracy, DR. Nav is also orders of magnitude faster: inference takes only 0.033 s per image (30 Hz), compared to 2–2.5 s per image (0.4–0.5 Hz) for the VLM baselines. This large speed gap highlights that our method is not only more accurate, but also computationally efficient enough for real-time navigation, whereas general-purpose VLMs are too slow to deploy on-board a robot. Together, these results demonstrate the clear benefit of lightweight RGB–LiDAR fusion over heavyweight, image-only VLMs for the task of dead-end prediction and recovery. This comparison highlights the benefit of combining geometric and semantic cues for early dead-end prediction and safe recovery.

\section{Conclusion, Limitations, \& Future work}
We present DR. Nav, a proactive dead-end detection and recovery framework that fuses RGB and LiDAR into a continuous semantic cost map. By predicting dead-end probabilities in real time and embedding them into the planning layer, our method demonstrated stronger anticipation of blocked regions compared to standard reactive planners and vision–language models. These results highlight the potential of proactive, semantics-aware navigation to reduce unsafe commitments and improve long-term reliability in challenging indoor and outdoor environments. However, the current study has important limitations. First, all experiments were conducted using replayed rosbag datasets, which provide repeatable evaluation but do not capture the full dynamics of real robots, including actuator delays, closed-loop feedback, and sensor degradation. Second, while our framework includes a recoverability module and a mechanism to rank recovery points, these components were not fully evaluated in the present results, leaving their effectiveness to be validated in physical deployments. Finally, the reliance on synchronized and noise-free sensor streams in the rosbag limits the realism of the evaluation, as real-world conditions often involve drift, occlusions, and hardware failures.

Future work will focus on addressing these limitations through real-time deployment on the Ghost quadruped robot. Key directions include (i) optimizing inference for on-board execution under strict latency constraints, (ii) robust synchronization of multi-sensor inputs and handling of missing or degraded data, (iii) extending the cost map to incorporate continuous motion feedback, slip conditions, and terrain interactions, and (iv) fully integrating and evaluating the recoverability framework, including ranking recovery points by their accessibility, energy cost, and long-term safety. This step is essential to demonstrate the unique contribution of proactive recoverability, ensuring that the robot not only avoids dead-ends but also selects the most efficient and reliable escape strategies when confronted with unavoidable blockages.


\bibliography{references}

\end{document}